\pdfoutput=1

\documentclass[11pt]{article}

\usepackage[preprint]{acl}

\usepackage{url}
\usepackage{times}
\usepackage{latexsym}
\usepackage{algorithm}
\usepackage{algorithmic}
\usepackage{booktabs} 
\usepackage{array}    
\usepackage{multirow}
\usepackage{fancyvrb}
\usepackage{xcolor}
\usepackage{svg}
\usepackage{enumitem}
\usepackage{tcolorbox}
\newtcolorbox{ChairmanPrompt}{
    colback=white,    
    colframe=black,        
    fonttitle=\bfseries,   
    title=Chairman Prompt, 
    boxrule=0.8mm,         
    sharp corners,         
    width=\textwidth,      
}

\newtcolorbox{CandidatePrompt}{
    colback=white,    
    colframe=black,        
    fonttitle=\bfseries,   
    title=Candidate Prompt, 
    boxrule=0.8mm,         
    sharp corners,         
    width=\textwidth,      
}

\newtcolorbox{ReviewerPrompt}{
    colback=white,    
    colframe=black,        
    fonttitle=\bfseries,   
    title=Reviewer Prompt, 
    boxrule=0.8mm,         
    sharp corners,         
    width=\textwidth,      
}

\newtcolorbox{EvaluationPrompt}{
    colback=white,    
    colframe=black,        
    fonttitle=\bfseries,   
    title=Evaluation Prompt, 
    boxrule=0.8mm,         
    sharp corners,         
    width=\textwidth,      
}

\usepackage{listings}
\usepackage{xcolor}

\lstdefinestyle{python1}{
    language=Python,
    basicstyle=\ttfamily\small,     
    commentstyle=\color{gray},      
    keywordstyle=\color{blue},      
    stringstyle=\color{red},        
    numbers=none,                   
}

\usepackage[T1]{fontenc}

\usepackage[utf8]{inputenc}

\usepackage{microtype}

\usepackage{inconsolata}

\usepackage{graphicx}

\title{Review-Instruct: A Review-Driven Multi-Turn Conversations Generation Method for Large Language Models}

\author{
 \textbf{Jiangxu Wu\textsuperscript},
 \textbf{Cong Wang\textsuperscript},
 \textbf{TianHuang Su\textsuperscript},
 \textbf{Jun Yang\textsuperscript},
\\
 \textbf{Haozhi Lin\textsuperscript},
 \textbf{Chao Zhang\textsuperscript},
 \textbf{Ming Peng\textsuperscript},
 \textbf{Kai Shi \textsuperscript},
\\
 \textbf{SongPan Yang\textsuperscript},
 \textbf{BinQing Pan\textsuperscript},
 \textbf{ZiXian Li\textsuperscript},
\\
\\
 \textsuperscript{}OPPO AI Center,
\\
  \small{
   \href{mailto:email@domain}{wujx27@mail2.sysu.edu.cn}
 }
}

\begin{document}
\pagestyle{empty}  
\maketitle
\begin{abstract}
The effectiveness of large language models (LLMs) in conversational AI is hindered by their reliance on single-turn supervised fine-tuning (SFT) data, which limits contextual coherence in multi-turn dialogues. Existing methods for generating multi-turn dialogue data struggle to ensure both diversity and quality in instructions. To address this, we propose Review-Instruct, a novel framework that synthesizes multi-turn conversations through an iterative "Ask-Respond-Review" process involving three agent roles: a Candidate, multiple Reviewers, and a Chairman. The framework iteratively refines instructions by incorporating Reviewer feedback, enhancing dialogue diversity and difficulty. We construct a multi-turn dataset using the Alpaca dataset and fine-tune the LLaMA2-13B model. Evaluations on MT-Bench, MMLU-Pro, and Auto-Arena demonstrate significant improvements, achieving absolute gains of 2.9\% on MMLU-Pro and 2\% on MT-Bench compared to prior state-of-the-art models based on LLaMA2-13B. Ablation studies confirm the critical role of the Review stage and the use of multiple Reviewers in boosting instruction diversity and difficulty. Our work highlights the potential of review-driven, multi-agent frameworks for generating high-quality conversational data at scale. The source code are available at: \href{https://github.com/wjx-git/Review-Instruct}{this https URL.}
\end{abstract}

\section{Introduction}

The rapid advancement of large language models (LLMs)  
\citep{Han:2021,OpenAI:2022} has revolutionized conversational AI systems. While these models demonstrate impressive capabilities in single-turn interactions, their performance in multi-turn dialogues remains suboptimal due to limitations in existing supervised fine-tuning (SFT) data 
 \citep{kong2024platolm}. Current approaches predominantly rely on single-turn conversational data for SFT \citep{li-etal-2024-quantity,YangXu2023LIMA,Chen2023AlpaGasus}, which fundamentally constrains the models' ability to maintain contextual coherence in extended dialogues \citep{sun-etal-2024-parrot}.

This limitation stems from two primary challenges in multi-turn data acquisition. First, simple concatenation of single-turn dialogues produces artificial conversations lacking natural flow \citep{kong2024platolm}. Second, while human-AI interaction logs provide authentic data \citep{ChatbotArena,vicuna2023}, they raise privacy concerns and require substantial collection efforts. Recent attempts to automate dialogue generation through dual-agent "Ask-Respond" frameworks \citep{ding2023enhancing,xu2023baize} have shown promise but face inherent limitations. Specifically, the asymmetric capabilities of LLMs in question generation versus answering, coupled with data scarcity for training specialized query generators \citep{kong2024platolm}, hinder the production of high-quality multi-turn dialogues at scale.

\begin{figure*}[t]
  \includegraphics[width=1\linewidth]{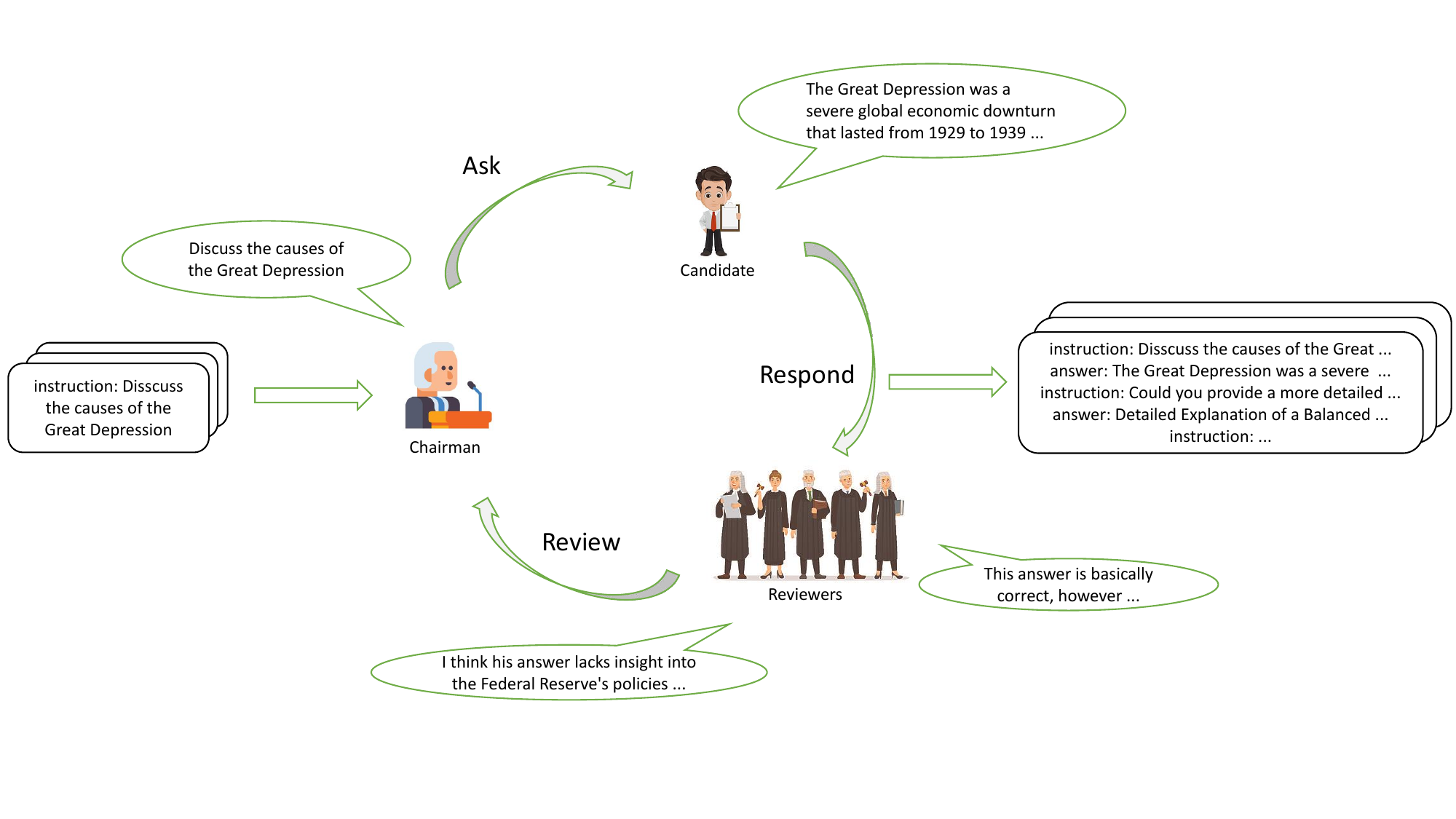} 
  \caption {Review-Instruct Iteration Process: the chairman questions the candidate, the candidate answers, and the reviewers evaluate those answers.}
  \label{fig:Review-Instruct}
\end{figure*}

To address these challenges, we propose Review-Instruct, a novel framework that introduces structured feedback mechanisms into dialogue generation. Drawing inspiration from the evaluation methodologies in Auto-Arena \citep{zhao2023autoarena}, our approach extends the conventional "Ask-Respond" paradigm with a critical Review stage, creating an "Ask-Respond-Review" pipeline. As illustrated in Figure~\ref{fig:Review-Instruct}, the framework employs three distinct AI agents: a Candidate, multiple Reviewers, and a Chairman. The process begins with the Chairman selecting an instruction from a predefined instruction dataset. Following the Candidate's response, multiple Reviewers conduct parallel evaluations using criteria including relevance, coherence, and depth. The Chairman then aggregates these assessments to generate contextually appropriate follow-up instructions, creating an iterative refinement loop that enhances both dialogue quality and complexity.

Our experimental results demonstrate significant improvements over existing approaches. Fine-tuning a LLaMA2-13B model\footnote{https://huggingface.co/meta-llama/Llama-2-13b-hf} with Review-Instruct generated data achieves state-of-the-art performance on key benchmarks: 2.9\% absolute improvement on MMLU-Pro \citep{wang2024mmlu}, 2.0\% absolute gain on MT-Bench \citep{zheng2023judging} and Superior performance in Auto-Arena pairwise comparisons \citep{zhao2023autoarena}. Ablation studies reveal two critical success factors: (1) The Review stage increases instruction diversity by 18.6\% and difficulty by 33.4\% compared to Ask-Respond paradigm, and (2) Multi-reviewer configurations generate 7.5\% more diversity instructions and 19.6\% more difficulty instructions than single-reviewer setups.

The principal contributions of this work are summarized as follows:
\begin{itemize}
\item [1)]
Novel Framework: Propose Review-Instruct, an "Ask-Respond-Review" pipeline integrating multi-agent feedback for high-quality multi-turn dialogue generation.
\item [2)]
Quality Enhancement Mechanism: The developed three-agent architecture (Candidate-Reviewers-Chairman) establishes an iterative refinement loop that automatically elevates dialogue complexity and difficulty.
\item [3)]
Empirical Validation: Comprehensive experiments establish new state-of-the-art results for 13B parameter models based on Llama2, with significant improvements across major benchmarks (2.9\% on MMLU-Pro, 2.0\% on MT-Bench). 
\end{itemize}

The remainder of this paper is organized as follows: Section 2 details the Review-Instruct methodology. Section 3 presents experimental results and benchmark comparisons. Sections 4 and 5 analyze the critical role of the review stage through comprehensive data-driven investigations. Section 6 reviews related work in SFT data generation. Sections 7 and 8 discuss future research directions and examine potential limitations of our approach, respectively. 

\section{Method}

This chapter presents our proposed Review-Instruct method. Section 2.1 provides a detailed explanation of this process, while Section 2.2 outlines the responsibilities of each role.

\subsection{Ask-Respond-Review}
Figure~\ref{fig:Review-Instruct} illustrates the Review-Instruct framework, which simulates an interview involving three roles: Candidate, Chairman, and Reviewers. The process begins with the chairman selecting a predefined instruction and presenting it to the candidate, who formulates a response. Reviewers then independently evaluate this response, identifying specific deficiencies.
This marks the completion of one cycle in the Ask-Respond-Review loop, resulting in an instruction-response pair. Following this, the process enters an iterative phase where the chairman synthesizes the reviewers' feedback to generate a revised instruction. The candidate then responds again, and the reviewers provide additional feedback, perpetuating the Ask-Respond-Review cycle. This iterative approach effectively evolves the original instruction into a multi-turn dialogue. The pseudo-code for this process is outlined in algorithm~\ref{alg1}.


\begin{algorithm}
\caption{Ask-Respond-Review cycle} 
\label{alg1} 
\begin{algorithmic}
\STATE data = queue() 
\STATE N = INPUT()
\FOR{Q in instructions}
        \STATE d = queue() 
        \FOR{n in range(1, N)}
        \STATE A = Respond(Q) 
        \STATE d.enqueue((Q, A)) 
        \STATE R = Review(A)
        \STATE Q = Ask(R)
        \ENDFOR
\STATE data.enqueue(d)
\ENDFOR
\end{algorithmic} 
\end{algorithm}


\subsection{Role Introduction}
The process involves three distinct roles: Chairman, Candidate, and Reviewers.

The Chairman is responsible for synthesizing follow-up instructions based on reviewers' feedback. The Chairman can dynamically utilize either a breadth-first or depth-first evolutionary strategy to generate new instructions.
Positive feedback triggers breadth-evolution, expanding the topic's scope to enhance instruction diversity.
Negative feedback triggers depth-evolution, focusing on specific weaknesses in the Candidate's prior response to elevate instruction difficulty. 
Examples of instruction evolution are provided in Appendix~\ref{sec:appendixB}.

The Candidate is responsible for generating a response to the Chairman's instruction.

Finally, a panel of independent Reviewers assesses the Candidate's response. 
This independent evaluation is crucial for mitigating potential bias and ensuring a diverse range of perspectives. 
Each Reviewer meticulously identifies any flaws or weaknesses in the Candidate's response, providing critical feedback that informs the subsequent direction of instruction evolution. 

Detailed prompts for each participant are provided in Appendix~\ref{sec:appendixA}.

\section{Experiments}
\subsection{Experimental Setup}
\subsubsection{Implementation}

The Review-Instruct is designed to operate with both instruction-only and instruction-response datasets.
When presented with instruction-only data, the instruction functions as the chairman question, commencing the synthesis process at the "Respond" stage. 
Conversely, for instruction-response datasets, the pre-existing response is utilized as the candidate response, thus the synthesis process initiating from the "Review" stage.

We applied Review-Instruct to the Alpaca dataset, which consists of single-turn instruction-response sample.
Traditional Ask-Respond methods like UltraLM \cite{ding2023enhancing} and Vicuna \cite{vicuna2023} use ChatGPT \cite{OpenAI:2022} to synthesize data in both the Ask and Respond phases. However, due to budget constraints, we cannot employ ChatGPT as the Chairman (Ask) or Candidate (Respond). Referring to the Chatbot-arena benchmark \cite{ChatbotArena}, Qwen1.5-14B-Chat demonstrates comparable capabilities to GPT-3.5. Therefore, we use Qwen1.5-14B-Chat \cite{qwen1.5} as both the Chairman and Candidate to eliminate potential confounding effects from model capability differences. A panel of reviewers included Qwen-2.5-32B-Instruct \cite{qwen2.5}, Deepseek-2.5 \citep{deepseekv2}, and Llama-3.1-70B-Instruct \citep{Llama32024}. 
Two dialogue turns were synthesized from the original data. No filtering or post-processing was applied.

Finally, we trained our model using the pre-trained LLaMA2-13B. 
We used the AdamW optimizer with an initial learning rate of 2e-5, a maximum token limit of 4096, and a batch size of 4 per GPU. 
Training was conducted across 32 A800 GPUs using Deepspeed Zero-3 for 3 epochs.

\begin{table*}[ht]
\centering 
\begin{tabular}{c c c c c} 
\toprule  
Model & \multicolumn{3}{c}{MT-Bench} & MMLU-Pro \\
\cmidrule(lr){2-4} 
 & turn\_1 & turn\_2 & Overall &  \\
\midrule  
LLaMA-2-13B-Chat & 7.06 & 6.24 & 6.65 & 25.34\% \\
UltraLM-13b-v2.0 & 6.90 & 6.36 & 6.63 & 26.88\% \\
Vicuna-13b-v1.5 & 6.76 & 6.05 & 6.57 & 27.38\% \\
Parrot-13b & 7.18 & 6.90 & 7.04 & - \\
WizardLM-13B-V1.2 & \textbf{7.38} & 6.74 & 7.06 & 26.75\% \\
Review-Instruct-13b & 7.25 & \textbf{7.15} & \textbf{7.20} & \textbf{29.65\%} \\
\bottomrule 
\end{tabular}
\caption{\label{Table-1}
    The results on the MT-Bench and MMLU-Pro benchmarks. For the MT-Bench evaluation, we report scores for the first and second turns, as well as the average score. For the Review-Instruct-13b model, we report the average score across five independent runs.
  }
\end{table*}

\subsubsection{Baseline}

This section compares Review-Instruct with several state-of-the-art large language models (LLMs), all initialized from the Llama2-13b base model. 
These models differ primarily in their training data.
The compared models include:

\begin{itemize}
\item Llama2-13b-chat: Fine-tuned on 27,000 human-annotated instruction-tuning data points and further optimized with Reinforcement Learning from Human Feedback (RLHF) \citep{long2022rlhf}.
\item UltraLM-13b-v2.0: Trained on 1.5 million conversations from the UltraChat dataset. This dataset is constructed through iterative conversations between two ChatGPT APIs.
\item Vicuna-13b-v1.5: Trained on user-ChatGPT conversation logs collected from ShareGPT\footnote{https://sharegpt.com/}.
\item Parrot-13b: Trained using 40,000 conversations from the Parrot dataset, also constructed through iterative conversations leveraging ChatGPT and the Parrot-Ask model. 
\item  WizardLM-13B-V1.2: Trained using the Evol-Instruct \cite{xu2024wizardlm} dataset. 
\end{itemize}

\subsubsection{Benchmark}

MTBench \citep{zheng2023judging} introduces a benchmark specifically designed to evaluate the ability of language models to follow instructions over multiple turns of conversation. 
This benchmark leverages GPT-4 \citep{gpt4} to assess the quality of model responses, demonstrating a strong correlation between automated and human evaluations.

MMLU-Pro \citep{wang2024mmlu} is a new benchmark designed to assess the comprehension abilities of language models. 
It expands upon the existing MMLU \citep{hendryckstest2021} dataset by incorporating more demanding, reasoning-oriented questions.
Furthermore, MMLU-Pro increases the number of possible answers per question from four to ten.
This expansion significantly elevates the difficulty of the benchmark and minimizes the likelihood of achieving a correct answer by chance.

Auto-Arena \citep{zhao2023autoarena} is an innovative framework for automated LLM evaluation.
Leveraging LLM-powered agents, it orchestrates a multi-round "peer battle." First, an LLM examiner generates evaluation questions.
Two LLM candidates then respond, engaging in a back-and-forth designed to reveal performance differences.
Finally, an LLM judge committee deliberates and selects a winner, mitigating bias and promoting fairness.  
Demonstrating a strong 92.14\% correlation with human preferences, Auto-Arena offers a promising automated alternative to traditional human evaluation platforms.
By employing a debate-style multi-turn dialogue format, Auto-Arena more effectively showcases a model's capabilities in multi-turn conversational settings.

\subsection{Experimental Results}
\subsubsection{Main Results}

Table~\ref{Table-1} details the performance of Review-Instruct-13b on the MT-Bench and MMLU-Pro benchmarks.The model achieves a superior overall MT-Bench score of 7.20, outperforming competing models across a range of tasks. The leading second-turn score of 7.15 is of particular interest, as it underscores the model's capabilities in multi-turn dialogue. In addition, Review-Instruct-13b achieves an accuracy of 29.65\% on the MMLU-Pro benchmark, a considerable improvement over other baseline models, demonstrating its proficiency in knowledge-intensive tasks.

\begin{figure}
  \includegraphics[width=\linewidth]{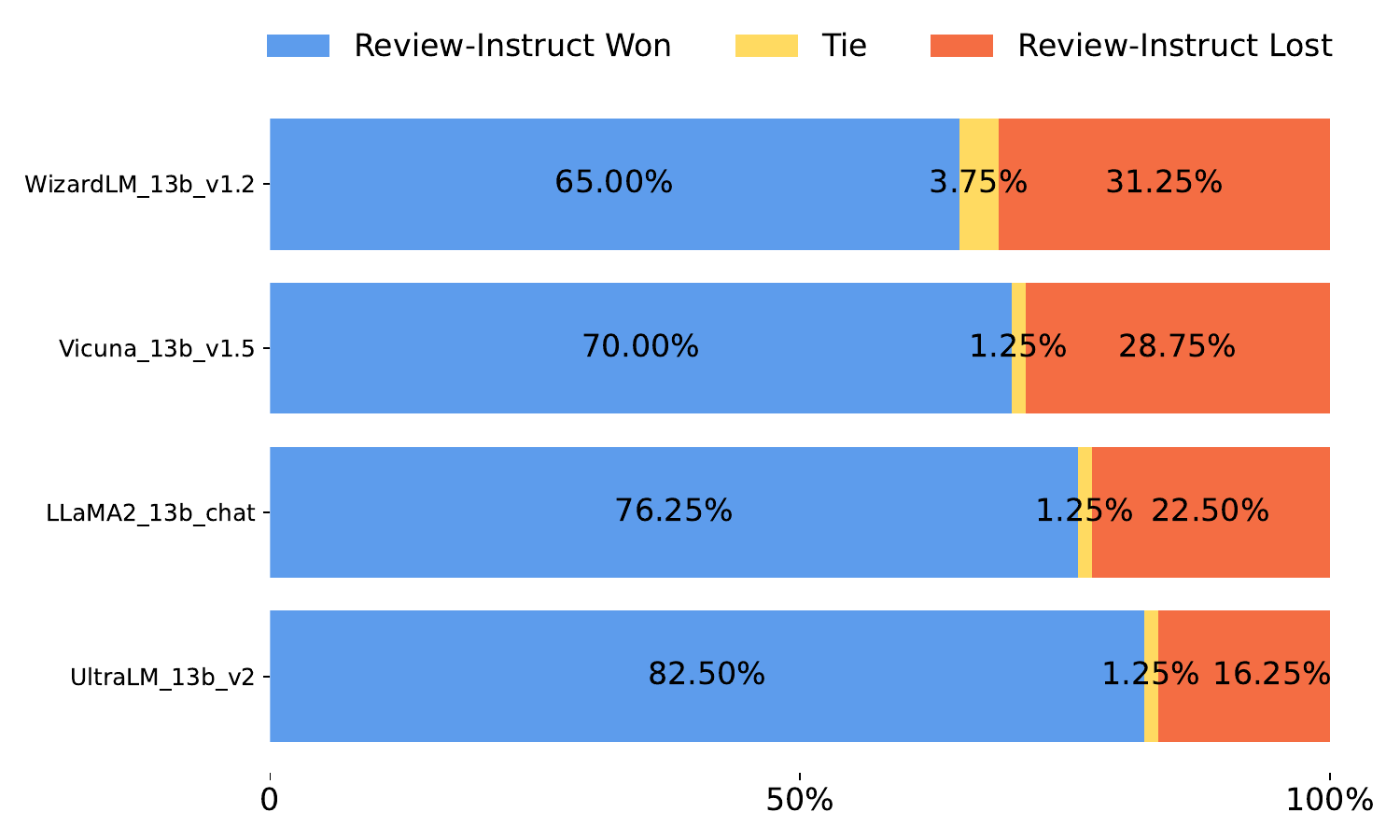} 
  \caption {We conducted evaluations using Auto-Arena, employing Review-Instruct and pairwise battles between all baseline models. GPT-4o served as the judge for these comparisons}
  \label{fig:auto-arena}
\end{figure}

Figure~\ref{fig:auto-arena} presents the performance of Review-Instruct-13b on the Auto-Arena benchmark. A direct comparison with Parrot's models was not feasible because their models are not publicly available. Nevertheless, when evaluated using GPT-4o \citep{gpt4o} as a judge, our model demonstrated a significant performance advantage over the baseline models. This not only highlights the superior comprehensive capabilities of our model, but also underscores its proficiency in handling multi-turn conversational scenarios.

\begin{table}  
\centering 
\begin{tabular}{c c c c} 
\toprule  
Ours vs & Won & Tie & Lost  \\
\midrule  
WizardLM\_13b\_v1.2 & 58 & 6 & 16  \\
Vicuna\_13b\_v1.5 & 53 & 7 & 12  \\
LLaMA2\_13b\_chat & 39 & 10 & 31  \\
UltraLM\_13b\_v2 & 62 & 4 & 14 \\
\bottomrule 

\end{tabular}
\caption{\label{Table-2}
    The results of the Manual Pair-Wise Evaluations.}
\end{table}

\begin{table*}[t]  
\centering 
\begin{tabular}{c c c c c} 
\toprule  
Model & \multicolumn{3}{c}{MT-Bench} & MMLU-Pro \\
\cmidrule(lr){2-4} 
 & turn\_1 & turn\_2 & Overall &  \\
\midrule  
Review-Instruct-single-turn & 6.89 & 5.91 & 6.4 & 19.3 \% \\
Review-Instruct-wo-review & 6.87 & 6.17 & 6.52 & 22.9 \% \\
Review-Instruct-one-review & 7.21 & 7.05 & 7.13 & 26.7 \% \\
Review-Instruct-13b & \textbf{7.25} & \textbf{7.15} & \textbf{7.20} & \textbf{29.65\%} \\
\bottomrule 

\end{tabular}
\caption{\label{Table-3}
    The results of our models trained on different datasets. Review-Instruct-single-turn refers to the model trained using a 156k dataset of single-turn conversations. Review-Instruct-wo-Review refers to the model trained on a multi-turn conversations, but with the Review stage removed. Review-Instruct-one-review means there is only one reviewer in the review stage. Review-Instruct represents the method proposed in this paper.
  }
\end{table*}

\begin{figure*}
  \includegraphics[width=1\linewidth]{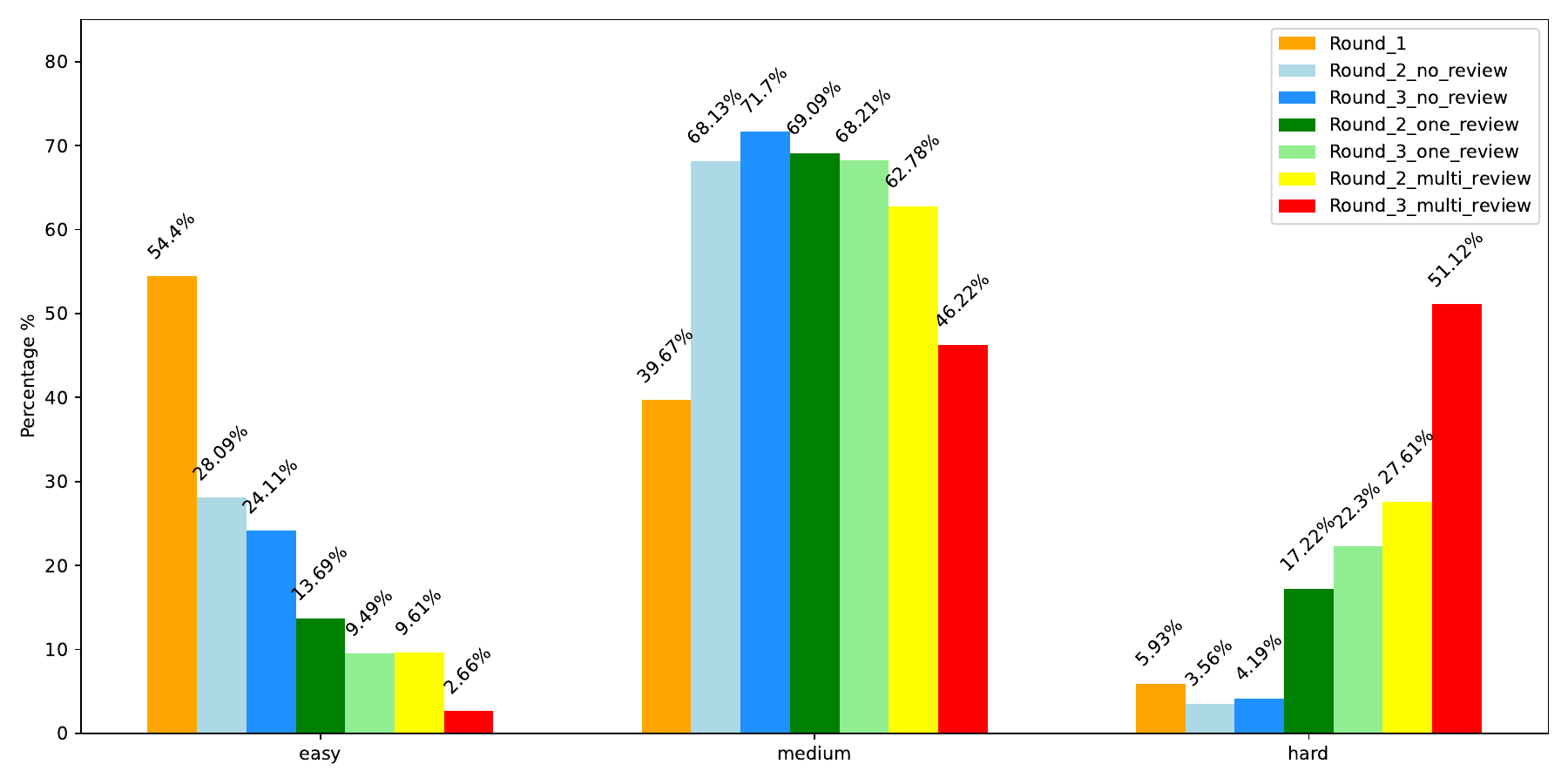}
  \caption {Instruction difficulty scores. Round 1 represents the original Alpaca data, serving as the foundation for subsequent synthesis.  Round 2 and 3 show the composition of newly added labels after the first and second rounds of synthetic data generation, respectively.\textit{no\_review} (no review stage), \textit{one\_review} (a single reviewer in the review stage), and \textit{multi\_review} (multiple reviewers in the review stage).  }
  \label{fig4}
\end{figure*}

We randomly selected 80 data entries from the auto-arena evaluation dataset, and we recruited four volunteers as evaluation experts. The results of the Manual Pair-Wise Evaluations are in table~\ref{Table-2}. From the manual evaluation results, our model significantly outperforms the baseline model in multi-turn dialogue scenarios.

The consistent superior performance of Review-Instruct-13b across the MT-Bench, MMLU-Pro, and Auto-Arena benchmarks strongly demonstrates its capabilities. 
Importantly, Review-Instruct's ability to create multi-turn dialogues from existing instructions datasets offers a simple, effective way to enhance datasets and improve model performance, demonstrating broad applicability.

\subsubsection{Ablation Study}
Three comparative experiments were designed to isolate and analyze the individual contributions of the following: (1) multi-turn versus single-turn conversations, (2) the impact of the Review stage, and (3) the influence of reviewer count within the Review stage.

To ensure that the observed benefits of multi-turn conversations were not merely a consequence of increased token count, we conducted a comparative analysis of models trained on multi-turn conversations and models trained on single-turn conversations with a controlled, equivalent number of tokens. 
The single-turn conversation dataset was constructed by extracting individual turns from the 52000 Review-Instruct multi-turn conversations and rephrasing them to eliminate contextual dependencies. This yielded a dataset of 156,000 single-turn conversations. 
Table~\ref{Table-3} reveals that the model trained on the single-turn conversations exhibited significantly lower performance on both benchmarks relative to the model trained on the multi-turn conversations. 
This finding underscores the importance of multi-turn conversations for achieving superior conversational coherence and knowledge utilization. 
Furthermore, the observed performance disparity cannot be solely attributed to the greater number of tokens present in the multi-turn conversations.

To isolate the impact of the Review phase, we created "Review-Instruct-wo-Review," a modified version of the Review-Instruct dataset that omits the Review process. In this variant, follow-up questions were generated directly from the candidate's responses. As shown in Table~\ref{Table-3}, Review-Instruct consistently achieved higher performance than Review-Instruct-wo-Review. This result underscores the crucial role of the Review phase in enhancing model performance.

Finally, a comparison was made between the efficacy of single- and multi-reviewer.
Table~\ref{Table-3} details the performance of "Review-Instruct-one-review" (single-reviewer) and "Review-Instruct" (multi-reviewer). The former exhibited a minor decrement in performance on the MT-Bench, but a considerably larger reduction on the MMLU-Pro benchmark. 
This disparity suggests that the multi-reviewer approach is more conducive to realizing the model's complete knowledge potential.

\section{Dataset Analysis}
As previously mentioned, the common method for synthesizing multi-turn conversations relies on the Ask-Respond paradigm.
Building upon this foundation, our proposed Review-Instruct framework introduces a novel Review stage.
This section provides a data-driven analysis of the Review stage's impact, specifically examining its influence on instruction diversity and difficulty. These metrics are crucial for assessing the quality of instruction-tuning data\citep{Llama32024,qwen2.5,long-etal-2024-llms}.

\begin{figure}
  \includegraphics[width=1\linewidth]{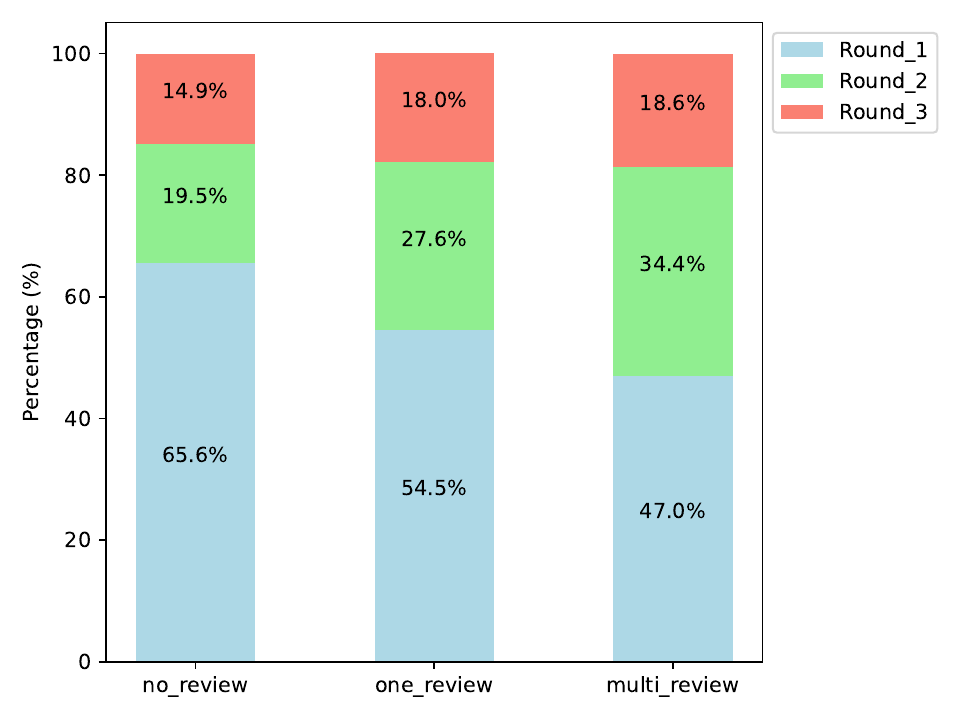} 
  \caption {Diversity metric. The proportion of newly added labels with each turn instruction relative to the total number of unique labels present in the entire dataset. The blue bars represent the percentage of tags present in the first round but not in the subsequent rounds, the green bars represent tags present in the second round but not in the first, and the red bars represent tags present in the third round but not in the first or second.}
  \label{fig3}
\end{figure}

\subsection{Instruction Difficulty}

The difficulty of instructions plays a significant role in the development of model reasoning abilities \citep{Llama32024}. 
Similar to Magpie \citep{zhang2024magpie} and llama3.1 \citep{Llama32024}, we used the gpt-4o \citep{gpt4o} to assess the difficulty of each instruction, categorizing them as "easy," "medium," or "hard." , specific prompts can be found in Appendix ~\ref{sec:appendixC}. 
Figure~\ref{fig4} presents the difficulty distribution for instructions generated by Review-Instruct and comparative methods. The blue bars represent the percentage of tags present in the first round but not in the subsequent rounds, the green bars represent tags present in the second round but not in the first, and the red bars represent tags present in the third round but not in the first or second.
The results demonstrate that the review stage significantly increases the proportion of "hard" instructions by 33.4\% compared to the no-review method and by 19.6\% compared to the single-review method.
This analysis confirms that the review stage enhances instruction difficulty, with the multi-reviewer approach providing the most significant improvement.

\subsection{Instruction Diversity}
Instruction diversity is crucial for ensuring model generalization. We adopt the diversity metric used by Instag \citep{lu2023instag}, which quantifies diversity based on the number of unique instruction labels. Specifically, we employ Instagger\footnote{https://huggingface.co/OFA-Sys/InsTagger} to tag each instruction in each round. Our diversity metric is the percentage of new labels introduced in each round, relative to the total number of unique labels in the entire dataset. As shown in Figure ~\ref{fig3}, the multi-review method achieves higher diversity by 18.6\% compared to the no-review method and consistently achieves higher diversity than the single-review method by 7.5\%, suggesting that the involvement of multiple reviewers leads to a more diverse set of instructions.

\begin{table}
    \noindent\hrulefill 
    
    \textbf{Chairman:} Summarize the events of the 1787 Constitutional Convention. \\
    \textbf{Candidate:} The 1787 Constitutional Convention was convened to amend the Articles of ... and the "Three-Fifths Compromise" regarding the issue of slavery. On September 17... \\
    \textbf{Chairman:} Please explain the long-term impact of the "Three-Fifths Compromise" on the U.S. Constitution and political system. 

    \noindent\hrulefill 
    \caption{\label{Table-4}An example of breadth evolution.}
\end{table}

\section{Case study}

We conclude with a case analysis illustrating how instructions evolve throughout the Review-Instruct process. Two primary methods of instruction evolution, breadth evolution and depth evolution, were discussed in evol-instruct \citep{xu2024wizardlm}. These methods utilize specifically crafted prompts to guide the evolutionary process. A similar phenomenon is observed in Review-Instruct, but our approach is based on adaptively selecting evolutionary directions based on reviewers' feedback. If the feedback is predominantly positive, the chairman might pose a question expanding on the topic, reflecting a breadth evolution. Conversely, if the feedback highlights deficiencies or errors, the chairman will pose targeted questions to address these criticisms, indicative of a depth evolution.

An example illustrating breadth evolution is in Table~\ref{Table-4}, with the complete case presented in Appendix~\ref{sec:appendixB}.
In this instance, the candidate's response displayed no shortcomings, prompting the chairman to select a knowledge point from the answer for further exploration through additional questioning.

\begin{table}
    \noindent\hrulefill 
    
    \textbf{Chairman:} Write a function to convert Fahrenheit to Celsius. \\
    \textbf{Candidate:} Here is a Python function that converts Fahrenheit to Celsius:\\
    \begin{lstlisting}[style=python1]
def fahrenheit_to_celsius(fahrenheit):
    celsius = (fahrenheit - 32) * 5/9
    return celsius
    \end{lstlisting}
    You can ... \\
    \textbf{Chairman:} How can this function be improved to handle non-numeric inputs and return a meaningful error message? 

    \noindent\hrulefill 

    \caption{\label{Table-5}An example of depth evolution.}
    
\end{table}

An example of depth evolution is provided in Table~\ref{Table-5}, also fully documented in Appendix~\ref{sec:appendixB}. 
This instruction involves creating a function to convert Fahrenheit to Celsius. 
The candidate's answer did not account for potential outlier input values, leading the chairman to require the implementation of anomaly handling, thereby increasing the depth of the question.

\section{Related Work}
The development of advanced multi-turn dialogue and instruction-following capabilities in LLMs relies heavily on quality supervised fine-tuning data.
While models such as Vicuna \citep{vicuna2023} and RealChat 
 \citep{zheng2024lmsyschatm} effectively utilize user-LLM interaction logs from platforms like ShareGPT and Chatbot Arena~\citep{ChatbotArena}, the manual creation of instructions is both labor-intensive and time-consuming, not to mention fraught with privacy concerns.
In light of recent advancements in LLM technology, there has been a surge in research focused on the automated generation of instruction datasets.

Various methods have emerged for this purpose.
For example, Self-Instruct \citep{selfinstruct} begins with a small collection of seed instructions and expands these using few-shot prompt.
Evol-Instruct \citep{xu2024wizardlm} refines or develops new instructions through a systematic process of in-depth and breadth evolution.
Humpback \citep{li2024selfalignment} uses instruction backtranslation to generate instructions based on web content.
Magpie \citep{zhang2024magpie} employs a self-synthesis approach, utilizing chat model to automate instruction generation. 
Persona-Hub \citep{personhub} introduces a persona-driven strategy, guiding LLMs to create instructions from diverse perspectives.
Although these methods demonstrate effectiveness in synthesizing a wide range of high-quality instructions, they typically focus on single-turn conversations, which limits their applicability in multi-turn conversations scenarios.

Some recent approaches have begun to explore the generation of multi-turn conversations.
Baize \citep{xu2023baize} and UltraChat \citep{ding2023enhancing}, for instance, leverage two ChatGPT APIs to simulate interactions between users and assistants. 
PlatoLM \citep{kong2024platolm} and Parrot \citep{sun-etal-2024-parrot} adopt a different strategy, training user simulators to interact with ChatGPT. These methods can all be categorized under the Ask-Respond pattern.

In contrast to existing approaches, our proposed Review-Instruct method augments the Ask-Respond paradigm with a novel Review stage. The incorporation of this Review stage, and the utilization of the reviews generated therein, facilitates the creation of instructions that demonstrate a marked improvement in both diversity and difficulty. Furthermore, the ability to synthesize multi-turn conversations from existing instruction datasets broadens the applicability of our approach. 
Our proposed method can be effectively integrated with existing approaches such as self-Instruct \citep{selfinstruct}, evol-Instruct \citep{xu2024wizardlm}, magpie \citep{zhang2024magpie}, and persona-hub \citep{personhub}. These methods can generate extensive instruction datasets, which can then be synthesized into multi-turn dialogue data using our approach.

\section{Conclusion and Future Work}
This study explores methods for synthesizing multi-turn conversational data.
We introduce the Review-Instruct, a novel approach based on the Ask-Respond-Review framework, which transforms instructions into multi-turn conversations.
Experimental results demonstrate that this method is of high quality and significantly enhances model performance.
Further analysis reveals that the Review process substantially increases both the diversity and difficulty of the instructions compared to a basic Ask-Respond pattern.

Beyond its capability in synthesizing textual instruction data, the proposed framework exhibits considerable potential for generating multi-modal instruction data. In multi-modal settings, the iterative process initiates with a single image presented to the chairman. The chairman then generates a question based on this visual stimulus, which the candidate is required to address. The remaining steps of the procedure align with the methodology outlined in the preceding sections.
A case illustrating is provided in Appendix~\ref{sec:appendixD}, highlighting directions for our future research. 

\section{Limitations}
The method proposed in this paper has the following limitations:

\textit{Dependency on Existing Instruction Datasets}: The Review-Instruct relies on a predefined instruction dataset for initialization. This means that the difficulty and diversity of the generated multi-turn conversations are, to some extent, constrained by the initial dataset.

\textit{Complexity of the Synthesis Process}: 
The complexity of the Review stage may also lead to increased time and computational costs for generating conversations. This could be a problem in resource-constrained environments.



\textit{Quality of LLMs}: The entire process hinges on the capabilities of the LLMs. If these LLMs are not strong enough, the reviews might be superficial, unhelpful, or even misleading. The Chairman's ability to synthesize reviews and generate a good follow-up instruction is also crucial.

\textit{Potential for Bias Amplification}: If the initial instruction dataset or the LLMs have biases, the Review-Instruct process could amplify these biases. For example, if the Reviewers consistently favor a particular style or viewpoint, the generated conversations could become increasingly skewed.

\textit{Generalizability Beyond Alpaca}: The experiments are based on the Alpaca dataset. It's not clear how well Review-Instruct would generalize to other types of instruction datasets or conversational domains.

\bibliography{acl_latex}

\begin{thebibliography}{30}
\providecommand{\natexlab}[1]{#1}

\bibitem[{Chiang et~al.(2023)Chiang, Li, Lin, Sheng, Wu, Zhang, Zheng, Zhuang, Zhuang, Gonzalez, Stoica, and Xing}]{vicuna2023}
Wei-Lin Chiang, Zhuohan Li, Zi~Lin, Ying Sheng, Zhanghao Wu, Hao Zhang, Lianmin Zheng, Siyuan Zhuang, Yonghao Zhuang, Joseph~E. Gonzalez, Ion Stoica, and Eric~P. Xing. 2023.
\newblock \href {https://lmsys.org/blog/2023-03-30-vicuna/} {Vicuna: An open-source chatbot impressing gpt-4 with 90\%* chatgpt quality}.

\bibitem[{Chiang et~al.(2024)Chiang, Zheng, Sheng, Angelopoulos, Li, Li, Zhang, Zhu, Jordan, Gonzalez, and Stoica}]{ChatbotArena}
Wei-Lin Chiang, Lianmin Zheng, Ying Sheng, Anastasios~Nikolas Angelopoulos, Tianle Li, Dacheng Li, Hao Zhang, Banghua Zhu, Michael Jordan, Joseph~E. Gonzalez, and Ion Stoica. 2024.
\newblock \href {https://arxiv.org/abs/2403.04132} {Chatbot arena: An open platform for evaluating llms by human preference}.
\newblock \emph{Preprint}, arXiv:2403.04132.

\bibitem[{DeepSeek-AI(2024)}]{deepseekv2}
DeepSeek-AI. 2024.
\newblock \href {https://arxiv.org/abs/2405.04434} {Deepseek-v2: A strong, economical, and efficient mixture-of-experts language model}.
\newblock \emph{Preprint}, arXiv:2405.04434.

\bibitem[{Ding et~al.(2023)Ding, Chen, Xu, Qin, Zheng, Hu, Liu, Sun, and Zhou}]{ding2023enhancing}
Ning Ding, Yulin Chen, Bokai Xu, Yujia Qin, Zhi Zheng, Shengding Hu, Zhiyuan Liu, Maosong Sun, and Bowen Zhou. 2023.
\newblock \href {https://arxiv.org/abs/2305.14233} {Enhancing chat language models by scaling high-quality instructional conversations}.
\newblock \emph{arXiv preprint arXiv:2305.14233}.

\bibitem[{Ge et~al.(2024)Ge, Chan, Wang, Yu, Mi, and Yu}]{personhub}
Tao Ge, Xin Chan, Xiaoyang Wang, Dian Yu, Haitao Mi, and Dong Yu. 2024.
\newblock \href {https://arxiv.org/abs/2406.20094} {Scaling synthetic data creation with 1,000,000,000 personas}.

\bibitem[{Han and Zhang(2021)}]{Han:2021}
Xu~Han and Zhengyan Zhang. 2021.
\newblock \href {https://doi.org/10.1016/j.aiopen.2021.08.002} {Pre-trained models: Past, present and future.}
\newblock \emph{AI Open}, 2:225--250.

\bibitem[{Hendrycks et~al.(2021)Hendrycks, Burns, Basart, Zou, Mazeika, Song, and Steinhardt}]{hendryckstest2021}
Dan Hendrycks, Collin Burns, Steven Basart, Andy Zou, Mantas Mazeika, Dawn Song, and Jacob Steinhardt. 2021.
\newblock Measuring massive multitask language understanding.
\newblock \emph{Proceedings of the International Conference on Learning Representations (ICLR)}.

\bibitem[{Kong et~al.(2024)Kong, Fan, Wan, Jiang, and Wang}]{kong2024platolm}
Chuyi Kong, Yaxin Fan, Xiang Wan, Feng Jiang, and Benyou Wang. 2024.
\newblock \href {https://arxiv.org/abs/2308.11534} {Platolm: Teaching llms in multi-round dialogue via a user simulator}.
\newblock In \emph{Proceedings of the 62nd Annual Meeting of the Association for Computational Linguistics (Volume 1: Long Papers)}, pages 7841--7863.

\bibitem[{Li et~al.(2024{\natexlab{a}})Li, Zhang, Li, Chen, Chen, Cheng, Wang, Zhou, and Xiao}]{li-etal-2024-quantity}
Ming Li, Yong Zhang, Zhitao Li, Jiuhai Chen, Lichang Chen, Ning Cheng, Jianzong Wang, Tianyi Zhou, and Jing Xiao. 2024{\natexlab{a}}.
\newblock \href {https://aclanthology.org/2024.naacl-long.421} {From quantity to quality: Boosting {LLM} performance with self-guided data selection for instruction tuning}.
\newblock In \emph{Proceedings of the 2024 Conference of the North American Chapter of the Association for Computational Linguistics: Human Language Technologies (Volume 1: Long Papers)}, pages 7595--7628, Mexico City, Mexico. Association for Computational Linguistics.

\bibitem[{Li et~al.(2024{\natexlab{b}})Li, Yu, Zhou, Schick, Levy, Zettlemoyer, Weston, and Lewis}]{li2024selfalignment}
Xian Li, Ping Yu, Chunting Zhou, Timo Schick, Omer Levy, Luke Zettlemoyer, Jason~E Weston, and Mike Lewis. 2024{\natexlab{b}}.
\newblock \href {https://openreview.net/forum?id=1oijHJBRsT} {Self-alignment with instruction backtranslation}.
\newblock In \emph{The Twelfth International Conference on Learning Representations}.

\bibitem[{Lichang et~al.(2023)Lichang, Shiyang, Jun, Hai, Kalpa, Vikas, Zheng, Vijay, Tianyi, Heng, and Hongxia}]{Chen2023AlpaGasus}
Chen Lichang, Li~Shiyang, Yan Jun, Wang Hai, Gunaratna Kalpa, Yadav Vikas, Tang Zheng, Srinivasan Vijay, Zhou Tianyi, Huang Heng, and Jin Hongxia. 2023.
\newblock \href {https://arxiv.org/abs/2307.08701} {Alpagasus: Train a better alpaca with fewer data.}
\newblock \emph{arXiv preprint arXiv:2307.08701}.

\bibitem[{Long et~al.(2024)Long, Wang, Xiao, Zhao, Ding, Chen, and Wang}]{long-etal-2024-llms}
Lin Long, Rui Wang, Ruixuan Xiao, Junbo Zhao, Xiao Ding, Gang Chen, and Haobo Wang. 2024.
\newblock \href {https://doi.org/10.18653/v1/2024.findings-acl.658} {On {LLM}s-driven synthetic data generation, curation, and evaluation: A survey}.
\newblock In \emph{Findings of the Association for Computational Linguistics: ACL 2024}, pages 11065--11082, Bangkok, Thailand. Association for Computational Linguistics.

\bibitem[{Lu et~al.(2023)Lu, Yuan, Yuan, Lin, Lin, Tan, Zhou, and Zhou}]{lu2023instag}
Keming Lu, Hongyi Yuan, Zheng Yuan, Runji Lin, Junyang Lin, Chuanqi Tan, Chang Zhou, and Jingren Zhou. 2023.
\newblock \href {https://arxiv.org/abs/2308.07074} {Instag: Instruction tagging for analyzing supervised fine-tuning of large language models}.
\newblock \emph{Preprint}, arXiv:2308.07074.

\bibitem[{{Meta}(2024)}]{Llama32024}
{Meta}. 2024.
\newblock \href {https://ai.meta.com/blog/meta-llama-3-1/} {Introducing llama 3.1: Our most capable models to date}.

\bibitem[{{OpenAI}(2022)}]{OpenAI:2022}
{OpenAI}. 2022.
\newblock Chatgpt: Optimizing language models for dialogue.
\newblock \url{https://chatgpt.r4wand.eu.org/}.

\bibitem[{{OpenAI}(2024)}]{gpt4o}
{OpenAI}. 2024.
\newblock \href {https://openai.com/index/hello-gpt-4o/} {Hello gpt-4o}.

\bibitem[{{OpenAI and Josh Achiam and Steven Adler and Sandhini Agarwal and Lama Ahmad and Ilge Akkaya and Florencia Leoni Aleman et al}(2023)}]{gpt4}
{OpenAI and Josh Achiam and Steven Adler and Sandhini Agarwal and Lama Ahmad and Ilge Akkaya and Florencia Leoni Aleman et al}. 2023.
\newblock \href {https://arxiv.org/abs/2303.08774} {Gpt-4 technical report}.

\bibitem[{Ouyang et~al.(2022)Ouyang, Wu, Jiang, Almeida, Wainwright, Mishkin, Zhang, Agarwal, Slama, Ray, Schulman, Hilton, Kelton, Miller, Simens, Askell, Welinder, Christiano, Leike, and Lowe}]{long2022rlhf}
Long Ouyang, Jeff Wu, Xu~Jiang, Diogo Almeida, Carroll~L. Wainwright, Pamela Mishkin, Chong Zhang, Sandhini Agarwal, Katarina Slama, Alex Ray, John Schulman, Jacob Hilton, Fraser Kelton, Luke Miller, Maddie Simens, Amanda Askell, Peter Welinder, Paul Christiano, Jan Leike, and Ryan Lowe. 2022.
\newblock \href {https://arxiv.org/abs/2203.02155} {Training language models to follow instructions with human feedback}.
\newblock \emph{arXiv preprint arXiv:2203.02155}.

\bibitem[{Sun et~al.(2024)Sun, Liu, Zhou, Huang, Song, Zhao, Zhang, Zhang, and Gai}]{sun-etal-2024-parrot}
Yuchong Sun, Che Liu, Kun Zhou, Jinwen Huang, Ruihua Song, Xin Zhao, Fuzheng Zhang, Di~Zhang, and Kun Gai. 2024.
\newblock \href {https://doi.org/10.18653/v1/2024.acl-long.525} {Parrot: Enhancing multi-turn instruction following for large language models}.
\newblock In \emph{Proceedings of the 62nd Annual Meeting of the Association for Computational Linguistics (Volume 1: Long Papers)}, pages 9729--9750, Bangkok, Thailand. Association for Computational Linguistics.

\bibitem[{Team(2024)}]{qwen1.5}
Qwen Team. 2024.
\newblock \href {https://qwenlm.github.io/blog/qwen1.5/} {Introducing qwen1.5}.

\bibitem[{Wang et~al.(2022)Wang, Kordi, Mishra, Liu, Smith, Khashabi, and Hajishirzi}]{selfinstruct}
Yizhong Wang, Yeganeh Kordi, Swaroop Mishra, Alisa Liu, Noah~A. Smith, Daniel Khashabi, and Hannaneh Hajishirzi. 2022.
\newblock \href {https://arxiv.org/abs/2212.10560} {Self-instruct: Aligning language model with self generated instructions}.

\bibitem[{Wang et~al.(2024)Wang, Ma, Zhang, Ni, Chandra, Guo, Ren, Arulraj, He, Jiang et~al.}]{wang2024mmlu}
Yubo Wang, Xueguang Ma, Ge~Zhang, Yuansheng Ni, Abhranil Chandra, Shiguang Guo, Weiming Ren, Aaran Arulraj, Xuan He, Ziyan Jiang, et~al. 2024.
\newblock \href {https://arxiv.org/abs/2406.01574} {Mmlu-pro: A more robust and challenging multi-task language understanding benchmark}.
\newblock \emph{arXiv preprint arXiv:2406.01574}.

\bibitem[{Xu et~al.(2024{\natexlab{a}})Xu, Sun, Zheng, Geng, Zhao, Feng, Tao, Lin, and Jiang}]{xu2024wizardlm}
Can Xu, Qingfeng Sun, Kai Zheng, Xiubo Geng, Pu~Zhao, Jiazhan Feng, Chongyang Tao, Qingwei Lin, and Daxin Jiang. 2024{\natexlab{a}}.
\newblock \href {https://openreview.net/forum?id=CfXh93NDgH} {Wizard{LM}: Empowering large pre-trained language models to follow complex instructions}.
\newblock In \emph{The Twelfth International Conference on Learning Representations}.

\bibitem[{Xu et~al.(2023)Xu, Guo, Duan, and McAuley}]{xu2023baize}
Canwen Xu, Daya Guo, Nan Duan, and Julian McAuley. 2023.
\newblock \href {https://arxiv.org/abs/2304.01196} {Baize: An open-source chat model with parameter-efficient tuning on self-chat data}.
\newblock \emph{arXiv preprint arXiv:2304.01196}.

\bibitem[{Xu et~al.(2024{\natexlab{b}})Xu, Jiang, Niu, Deng, Poovendran, Choi, and Lin}]{zhang2024magpie}
Zhangchen Xu, Fengqing Jiang, Luyao Niu, Yuntian Deng, Radha Poovendran, Yejin Choi, and Bill~Yuchen Lin. 2024{\natexlab{b}}.
\newblock \href {https://arxiv.org/abs/2406.08464} {Magpie: Alignment data synthesis from scratch by prompting aligned llms with nothing}.
\newblock \emph{arXiv preprint arXiv:2406.08464}.

\bibitem[{Yang et~al.(2024)Yang, Yang, Zhang, Hui, Zheng, Yu, Li, Liu, Huang, Wei, Lin, Yang, Tu, Zhang, Yang, Yang, Zhou, Lin, Dang, Lu, Bao, Yang, Yu, Li, Xue, Zhang, Zhu, Men, Lin, Li, Xia, Ren, Ren, Fan, Su, Zhang, Wan, Liu, Cui, Zhang, and Qiu}]{qwen2.5}
An~Yang, Baosong Yang, Beichen Zhang, Binyuan Hui, Bo~Zheng, Bowen Yu, Chengyuan Li, Dayiheng Liu, Fei Huang, Haoran Wei, Huan Lin, Jian Yang, Jianhong Tu, Jianwei Zhang, Jianxin Yang, Jiaxi Yang, Jingren Zhou, Junyang Lin, Kai Dang, Keming Lu, Keqin Bao, Kexin Yang, Le~Yu, Mei Li, Mingfeng Xue, Pei Zhang, Qin Zhu, Rui Men, Runji Lin, Tianhao Li, Tingyu Xia, Xingzhang Ren, Xuancheng Ren, Yang Fan, Yang Su, Yichang Zhang, Yu~Wan, Yuqiong Liu, Zeyu Cui, Zhenru Zhang, and Zihan Qiu. 2024.
\newblock \href {https://arxiv.org/abs/2412.15115} {Qwen2.5 technical report}.
\newblock \emph{arXiv preprint arXiv:2412.15115}.

\bibitem[{Yang et~al.(2023)Yang, Yongqiang, Yufan, Mengnan, Maoquan, Bin, and Neel}]{YangXu2023LIMA}
Xu~Yang, Yao Yongqiang, Huang Yufan, Qi~Mengnan, Wang Maoquan, Gu~Bin, and Sundaresan Neel. 2023.
\newblock \href {https://arxiv.org/abs/2312.11508} {Rethinking the instruction quality: Lift is what you need}.
\newblock \emph{arXiv preprint arXiv:2312.11508}.

\bibitem[{Zhao et~al.(2024)Zhao, Zhang, Chia, Xu, Zhao, and Bing}]{zhao2023autoarena}
Ruochen Zhao, Wenxuan Zhang, Yew~Ken Chia, Weiwen Xu, Deli Zhao, and Lidong Bing. 2024.
\newblock \href {https://arxiv.org/abs/2405.20267} {Auto-arena: Automating llm evaluations with agent peer battles and committee discussions}.
\newblock \emph{arXiv preprint arXiv:2405.20267}.

\bibitem[{Zheng et~al.(2024)Zheng, Chiang, Sheng, Li, Zhuang, Wu, Zhuang, Li, Lin, Xing, Gonzalez, Stoica, and Zhang}]{zheng2024lmsyschatm}
Lianmin Zheng, Wei-Lin Chiang, Ying Sheng, Tianle Li, Siyuan Zhuang, Zhanghao Wu, Yonghao Zhuang, Zhuohan Li, Zi~Lin, Eric Xing, Joseph~E. Gonzalez, Ion Stoica, and Hao Zhang. 2024.
\newblock \href {https://openreview.net/forum?id=BOfDKxfwt0} {{LMSYS}-chat-1m: A large-scale real-world {LLM} conversation dataset}.
\newblock In \emph{The Twelfth International Conference on Learning Representations}.

\bibitem[{Zheng et~al.(2023)Zheng, Chiang, Sheng, Zhuang, Wu, Zhuang, Lin, Li, Li, Xing, Zhang, Gonzalez, and Stoica}]{zheng2023judging}
Lianmin Zheng, Wei-Lin Chiang, Ying Sheng, Siyuan Zhuang, Zhanghao Wu, Yonghao Zhuang, Zi~Lin, Zhuohan Li, Dacheng Li, Eric.~P Xing, Hao Zhang, Joseph~E. Gonzalez, and Ion Stoica. 2023.
\newblock \href {https://arxiv.org/abs/2306.05685} {Judging llm-as-a-judge with mt-bench and chatbot arena}.
\newblock \emph{Preprint}, arXiv:2306.05685.

\end{thebibliography}
\clearpage

\appendix
\section{Appendices}
\subsection{Role Prompt}
\label{sec:appendixA}
\begin{ChairmanPrompt}
<|im\_start|>system\\
This is an interview setting, and you, as the chair of the interview panel, are conducting an in-person interview. The candidate will strive to answer the questions you pose, and the panel members will evaluate the candidate's responses based on usefulness, relevance, accuracy, depth, and creativity. Based on these evaluations, you need to pose new questions to deeply assess the candidate. Here are your action guidelines:\\
<think>: Gradually analyze each panel member's comments. This is hidden from the candidate. Reflect only when necessary, and keep it concise.\\
<ask>: Summarize these comments and pose a potential follow-up question to deeply assess the candidate's abilities. If most of the reviewers' comments are positive, please raise a related field question based on the dialogue topic. If most of the reviewers' comments suggest that the answers are insufficient or incorrect, please raise targeted questions based on these criticisms. The question should elicit a concise response and avoid excessive specificity or repetition. If no panel comments are provided, do not ask a question!\\
Strictly follow the action guidelines. Conduct the assessment in English unless otherwise necessary.\\
<|im\_start|>user\\
Give three tips for staying healthy.<|im\_end|>\\
<|im\_start|>assistant\\
1.Eat a balanced diet and make sure to include plenty of fruits and vegetables. \\
2. Exercise regularly to keep your body active and strong. \\
3. Get enough sleep and maintain a consistent sleep schedule.<|im\_end|>\\
<|im\_start|>user\\
Action guide: only include <ask>. Use <think> if needed.  Finish your whole response within 300 words, including <think>. ENCLOSE EACH ACTION IN ITS RESPECTIVE TAGS!\\
Comments from members of the committee:\\
Judge 1: While your tips are generally sound and commonly recommended, they could benefit from more specificity and context. For instance, when you mention eating a balanced diet, it would be helpful to specify what constitutes a balanced diet, such as including lean proteins, whole grains, and healthy fats. Additionally, for exercise, specifying the type, duration, and frequency (e.g., 30 minutes of moderate-intensity exercise at least five times a week) would provide clearer guidance. Lastly, for sleep, mentioning the recommended number of hours (e.g., 7-9 hours per night for adults) and the importance of a sleep-friendly environment (e.g., dark, quiet, and cool room) would enhance the practicality of your advice.\\
Judge 2: While your tips are generally sound, they lack depth and specificity. For instance, "eat a balanced diet" is a broad statement. You could elaborate on what constitutes a balanced diet, such as including lean proteins, whole grains, and healthy fats. Additionally, "exercise regularly" could be more informative by suggesting types of exercises (e.g., cardio, strength training, flexibility exercises) and frequency (e.g., 150 minutes of moderate aerobic activity per week). Lastly, "get enough sleep" is vague; you could specify the recommended hours of sleep for different age groups and the importance of sleep quality.<|im\_end|>\\
<|im\_start|>assistant
\end{ChairmanPrompt}

\clearpage

\begin{CandidatePrompt}
<|im\_start|>system\\
This is an interview setting, and you, as the candidate, are undergoing an in-person interview. The interviewer will randomly ask questions, and you need to strive to provide the best possible answers. The interviewer will evaluate your responses based on usefulness, relevance, accuracy, depth, and creativity, and will ask new questions based on your answers to deeply assess you. Here are your action guidelines:\\
<think>: Gradually think to analyze the question or plan your answer. This is hidden from the interviewer. Reflect only when necessary, and keep it concise.\\
<respond>: Respond to the user's input as accurately as possible.\\
Strictly follow the action guide.<|im\_end|>\\
<|im\_start|>user\\
Give three tips for staying healthy.<|im\_end|>\\
<|im\_start|>assistant\\
1.Eat a balanced diet and make sure to include plenty of fruits and vegetables. \\
2. Exercise regularly to keep your body active and strong. \\
3. Get enough sleep and maintain a consistent sleep schedule.<|im\_end|>\\
<|im\_start|>user\\
Action guide: only include <respond>. Use <think> if needed. Finish your whole response within 300 words, including <think>. ENCLOSE EACH ACTION IN ITS RESPECTIVE TAGS!\\
Opponent's Response or Question:\\
Could you provide a more detailed explanation of what constitutes a balanced diet, including specific food groups and their recommended proportions? Additionally, how would you tailor this advice for someone with specific dietary restrictions, such as a vegetarian or someone with a gluten intolerance?<|im\_end|>\\
<|im\_start|>assistant
\end{CandidatePrompt}

\clearpage

\begin{ReviewerPrompt}
<|im\_start|>system\\
This is an interview setting, where you are a member of the interview panel conducting a live interview. The candidate will strive to answer the questions, and you will evaluate the candidate's responses based on aspects such as usefulness, relevance, accuracy, depth, and creativity. Below is your action guide:\\
<think>: Gradually think through to analyze the problem or plan a response. This is hidden from the candidate. Only think when necessary, and keep it concise.\\
<criticize>: Criticize the inadequacies and flaws in the candidate's answers.\\
Strictly follow the action guidelines. Conduct the assessment in English unless otherwise necessary.<|im\_end|>\\
<|im\_start|>user\\
Dear Examiners, I am ready to answer the questions. Please proceed.<|im\_end|>\\
<|im\_start|>assistant\\
Give three tips for staying healthy.<|im\_end|>\\
<|im\_start|>user\\
Action guide: only include <criticize>. Use <think> if needed. Finish your whole response within 300 words, including <think>. ENCLOSE EACH ACTION IN ITS RESPECTIVE TAGS!\\
Opponent\'s Response or Question: 1.Eat a balanced diet and make sure to include plenty of fruits and vegetables. \\
2. Exercise regularly to keep your body active and strong. \\
3. Get enough sleep and maintain a consistent sleep schedule.<|im\_end|>\\
<|im\_start|>assistant
\end{ReviewerPrompt}

\newblock
\subsection{Evolution Case}
\label{sec:appendixB}

\begin{table*}
\noindent\hrulefill \\
\textbf{>>chairman}\\
Summarize the events of the 1787 Constitutional Convention.\\
\textbf{>>candidate}\\
The 1787 Constitutional Convention was convened to amend the Articles of Confederation but ultimately led to the birth of the United States Constitution. The convention was held in May in Philadelphia and included representatives from 13 states (except Rhode Island). During the meeting, the delegates engaged in fierce debates over the structure of the national government and the distribution of powers, ultimately reaching several important compromises, including the "Great Compromise" between large and small states (which established the Senate and the House of Representatives) and the "Three-Fifths Compromise" regarding the issue of slavery. On September 17, the delegates signed the Constitution of the United States, establishing principles such as federalism and the separation of powers, laying the foundation for the modern political system of the United States.\\
\textbf{>>chairman}\\
Please explain the long-term impact of the "Three-Fifths Compromise" on the U.S. Constitution and political system.\\
\textbf{>>candidate}\\
The "Three-Fifths Compromise" stipulated that when calculating the population of each state for the purpose of determining representation in the House of Representatives and the allocation of direct taxes, every five slaves would be counted as three individuals. This compromise had a long-term impact on the U.S. Constitution and political system: first, it temporarily eased tensions between the northern and southern states, allowing the Constitution to be ratified; second, it reinforced the position of slavery in the southern states, making the issue of slavery one of the central political topics in 19th-century America, ultimately leading to the Civil War; and finally, it affected the allocation of House seats, giving southern states more representation in the federal government and enabling them to hold an advantage in political decision-making. It was not until the ratification of the 13th Amendment in 1865 that the "Three-Fifths Compromise" was formally rendered ineffective.

\noindent\hrulefill 
\caption{Breadth evolution case}
\label{tab:Breadth evolution}
\end{table*}

\begin{table*}
\noindent\hrulefill \\
\textbf{>>chairman}\\
Write a function to convert Fahrenheit to Celsius.\\
\textbf{>>candidate}\\
Here is a Python function that converts Fahrenheit to Celsius:\\

    \begin{lstlisting}[style=python1]
def fahrenheit_to_celsius(fahrenheit):
    celsius = (fahrenheit - 32) * 5/9
    return celsius
    \end{lstlisting}
    
You can use this function by passing a Fahrenheit value as an argument, and it will return the corresponding Celsius value. For example, to convert 32 degrees Fahrenheit to Celsius, you can call \textit{fahrenheit\_to\_celsius(32)}, and the result will be 0 degrees Celsius.\\
\textbf{>>chairman}\\
How can this function be improved to handle non-numeric inputs and return a meaningful error message?\\
\textbf{>>candidate}\\
To improve this function to handle non-numeric inputs, we can add a type check within the function. If the input Fahrenheit value is not a numeric type, we will return an error message. Here is the revised function:\\
    \begin{lstlisting}[style=python1]
def fahrenheit_to_celsius(fahrenheit):
    celsius = (fahrenheit - 32) * 5/9
    return celsius
    \end{lstlisting}
Now, if you try to call this function with a non-numeric input, such as \textit{fahrenheit\_to\_celsius("abc")}, it will return the error message: 'Error: The input Fahrenheit value must be a numeric type'.

\hrulefill 
\caption{Breadth evolution case}
\label{tab:Breadth_evolution}
\end{table*}

\clearpage

\begin{figure*}[ht]
\centering
  \includegraphics[width=0.5\linewidth]{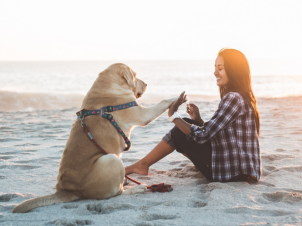} 
  \caption {The image for the multimodal case. The iterative Ask-Respond-Review process for generating synthetic multi-turn dialogues relies on different initiation mechanisms depending on the modality. For text-based dialogues, an initial instruction serves as the starting prompt, whereas for multi-modal dialogues, the process is initiated solely by an input image.}
  \label{fig:case}
\end{figure*}

\subsection{Evaluation Prompt}
\label{sec:appendixC}
\begin{EvaluationPrompt}
\# Instruction\\
You first need to identify the given user intent and then label the difficulty level of the user query based on the content of the user query. The user query is extracted from a conversation and may contain references, but this should not affect your judgment.\\
\#\# User Query\\
\{input\}\\
\#\# Output Format\\
Given the user query, in your output, you first need to identify the user intent and the knowledge needed to solve the task in the user query. Then, rate the difficulty level of the user query as 'easy', 'medium', or 'hard'.\\
Now, please output the user intent and difficulty level below in a json format by filling in the placeholders in [...]:\\
\{\{\\
"intent": "The user wants to [....]",\\
"knowledge": "To solve this problem, the models need to know [....]",\\
"difficulty": "[easy/medium/hard]"\\
\}\}
\end{EvaluationPrompt}

\subsection{Multi-Modal Case}
\label{sec:appendixD}

\begin{table*}[b]
\noindent\hrulefill \\
\textbf{>>chairman}\\
Please describe the interaction between the person and the dog in this picture, and discuss what you think this interaction means to both the person and the dog.\\
\textbf{>>candidate}\\
This picture shows a person interacting with a dog on a beach. The person and the dog are sitting face to face, and the dog is extending a paw to shake hands with the person, appearing very friendly and intimate.\\
This interaction has positive meanings for both the person and the dog. For the person, interacting with the dog can bring emotional satisfaction and happiness, help reduce stress and anxiety, and enhance the sense of well-being. For the dog, this interaction is not only an affirmation and reward for its behavior, but also strengthens its trust and dependence on its owner, promoting its psychological well-being.\\
Overall, this interaction between the person and the dog not only helps to build a deep emotional bond, but also enhances the quality of life for both.\\
\textbf{>>chairman}\\
What do you think this scene in the picture suggests about the relationship between humans and nature?\\
\textbf{>>candidate}\\
This picture shows a person interacting with a dog on a beach, presenting a beautiful scene of harmonious coexistence between humans and nature. Here's my understanding of the implications of this scene for the relationship between humans and nature:\\
1.Connecting with Nature: The person and the dog in the picture are enjoying the beautiful scenery of nature on the beach, which reminds us to spend more time connecting with nature and experiencing its tranquility and beauty. The natural environment can not only relax our body and mind, but also enhance our sense of well-being.\\
2.Harmonious Coexistence between Humans and Animals: The interaction between the person and the dog demonstrates the friendly relationship between humans and animals. This harmonious coexistence is not only beneficial to the psychological well-being of animals, but also allows humans to gain emotional satisfaction and happiness.\\
3.Awareness of Environmental Protection: In such a beautiful natural environment, we are more aware of the importance of protecting the environment. Only by protecting our natural environment can we and animals enjoy such wonderful moments.\\
4.Building Emotional Bonds: The interaction between the person and the dog is not just a simple game, but also the building of an emotional bond. This emotional bond is not limited to humans and animals, but can also extend to the relationship between humans and nature. We should cherish and maintain this bond to promote the harmonious development of humans and nature.\\
In summary, this picture inspires us to cherish the natural environment, treat animals kindly, and strive to build a harmonious ecological relationship, thereby enhancing our quality of life and well-being.

\noindent\hrulefill 
\caption{Multi-Modal Case. The input is the image shown in Figure ~\ref{fig:case}.}
\label{tab:Multi_Modal_Case}
\end{table*}

\end{document}